\documentclass[10pt, conference]{IEEEtran}
\IEEEoverridecommandlockouts

\usepackage[utf8]{inputenc}
\usepackage{graphicx}
\usepackage{booktabs}
\usepackage{array}
\usepackage{amsmath}
\usepackage{verbatim}
\usepackage{epsfig}
\usepackage{amssymb}
\usepackage{float}
\usepackage{tabularx}
\usepackage{bbm}
\usepackage{breqn}
\usepackage{booktabs}
\usepackage{placeins}

\usepackage[pagebackref=true,breaklinks=true,colorlinks,bookmarks=false]{hyperref}

\newcolumntype{L}[1]{>{\raggedright\arraybackslash}p{#1}}

\begin{document}
\title{Transfer Learning with Deep CNNs for Gender Recognition and Age Estimation}

\author{\IEEEauthorblockN{Philip Smith }
\IEEEauthorblockA{Computer Science\\
UNC Wilmington\\
North Carolina, United States\\
ps1994@uncw.edu}
\and
\IEEEauthorblockN{Cuixian Chen}
\IEEEauthorblockA{Mathematics and Statistics\\
UNC Wilmington\\
North Carolina, United States\\
chenc@uncw.edu}
}

\maketitle

\begin{abstract} 
In this project, competition-winning deep neural networks with pretrained weights are used for image-based gender recognition and age estimation. Transfer learning is explored using both VGG19 and VGGFace pretrained models by testing the effects of changes in various design schemes and training parameters in order to improve prediction accuracy. Training techniques such as input standardization, data augmentation, and label distribution age encoding are compared. Finally, a hierarchy of deep CNNs is tested that first classifies subjects by gender, and then uses separate male and female age models to predict age. A gender recognition accuracy of 98.7\% and an MAE of 4.1 years is achieved. This paper shows that, with proper training techniques, good results can be obtained by retasking existing convolutional filters towards a new purpose. 
\end{abstract}
\begin{IEEEkeywords}
transfer learning, deep learning, convolutional neural network, age estimation, gender classification 
\end{IEEEkeywords}

\IEEEpeerreviewmaketitle

\section{Introduction}
Deep learning has developed rapidly recently due to the availability and large scale of labeled data, and high performance computing. Deep learning is currently one of the most popular machine learning techniques in Artificial Intelligence (AI) \cite{gu2018recent, prieto2016neural, liu2017survey, guo2016deep, schmidhuber2015deep}. In computer vision, many classical techniques of feature extraction and subspace learning have been eclipsed due to the recent good performance in deep learning. 
Traditionally, good results have been obtained using combinations of classifiers, regressors, hand-crafted features, facial landmarks, and dimensionality reduction \cite{Cao:2008:GRB:1459359.1459470} \cite{GUO_joint_estimation_age_gender} \cite{classification_of_face_images}. Neural networks, however, have taken the scene with their ability to learn and memorize features, and keep improving in accuracy as more data are observed. Thus
Deep Neural Networks (DNNs) have far surpassed traditional classification and regression techniques, and have even surpassed human performance on a number of well-known benchmarks  \cite{han_otto_jain_2013} \cite{ILSVRC}.

\section{Related Works}
In a modern interconnected information society, it is critical to identify or verify individuals accurately at real-time. Due to its significant role in human computer interaction (HCI), internet access control, and security control and surveillance, face-based demographical research has attracted great attention in both research communities and industries \cite{fu2010age}. 
MORPH-II \cite{ricanek2006morph} has been the subject of many studies concerning age and gender estimation. As such, it is a good way to compare the efficacy of different techniques. Han et al. \cite{han_otto_jain_2013} gauges human age estimation by crowd-sourcing estimates on two popular face-image databases. They found estimates on the FG-NET dataset to be off by an average of 4.7 years. They mention that that number might be low because it is easy to guess the ages of babies and children without much variation in predictions. In fact, the average age error on FG-NET subjects older than 15 is 7.4 years, which is similar to the human error of 7.2 years that was calculated on the PSCO dataset. In the same study, Han et al. use a hierarchy of support vector machines (SVMs) and biologically-inspired features (BIFs) to obtain an average age estimation error of 4.2 years on the MORPH-II database. 

Deep learning 
is promising to allow for the full utilization of large datasets in order to solve machine learning problems. 
Amongst the different types of deep learning architectures, convolutional neural networks (CNN) have been proven to be very effective for human demographics estimation due to their proficiency at extracting precise details from images. 
Such studies include age estimation \cite{deep_expectation,hu2017facial,niu2016ordinal} and gender classification \cite{effective_training,castrillon2017descriptors,levi2015age} .
Niu et al. \cite{Niu_2016_CVPR} obtain an error of 3.28 years using ordinal regression CNNs and random splits of the MORPH-II dataset where 80\% of the images are used for training and 20\% are used for testing. 

Rothe et al. \cite{deep_expectation} considered deep CNNs for age classification problems. The VGG-16 architecture and IMDB-WIKI dataset are employed in this study. With a random split of 80\% for training and 20\% for testing on MORPH-II, it achieves a MAE of 2.68 with additional fine-tuning on IMDB-WIKI dataset before fine-tuning on MORPH-II dataset.   
Later, Antipov et al. \cite{antipov2017effective} extend the work from \cite{deep_expectation} and  
consider the problems of selection of optimal CNN architecture and training strategies. They conclude that Label Distribution Age Encoding (LDAE) \cite{geng2013facial} is an optimal way for the target encoding to train a CNN for an age estimation task. It is showed  that face recognition pretraining is more effective for deep gender and age CNNs comparing to general task pretraining. Following the subsetting scheme in \cite{guo2011simultaneous} for MORPH-II, it achieves a MAE of 2.99 years with VGG-16 pretrained CNN  for facial recognition, and a gender classification accuracy of 99.3\% with ResNet-50 pretrained CNN  for facial recognition. Their model also won the ChaLearn Apparent age estimation challenge in 2016 \cite{effective_training}.

%
In this paper, transfer learning is employed to tackle the problem of recognizing a person's age and gender from an image using deep CNNs. A variety of network designs and training techniques are explored. We consider dynamic LDAE, which outperforms the static LDAE considered in  \cite{antipov2017effective}. A gender-specified hierarchical age model is proposed in this study. Experimental results demonstrate its effectiveness over the general age model.

\section{Transfer Learning}
Because of the vastness and complexity of deep neural network architecture, designing and testing models is expensive and time-consuming. When approaching an AI problem, quick results can be obtained by utilizing a technique known as \textit{transfer learning}. In transfer learning, the weights and convolutional filters that are proficient at one task, can be reused for a different task requiring only a small amount of retraining. This involves using a network architecture with preloaded weights, modifying it slightly, and then retraining part or all of the model to output predictions for the new task. The filters learned by one task, such as classifying animals, are used to extract features from images that can then be interpreted by the retrained portion of the neural network in order to perform its new task. In this paper, the deep convolutional neural network known as VGG \cite{vgg_paper} is used to study transfer learning using two different types of pretraining.

\subsection{VGG19}
VGG19 \cite{vgg_paper} is a DNN architecture developed by Karen Simonyan and Andrew Zisserman of the Visual Geometry Group at Oxford. The ``19" in the name refers to the number of weight layers in the network. VGG16 was considered to be more successful in the ImageNet competition in 2014 and tied with GoogLeNet, however, the extra depth of VGG19 was leveraged to achieve better results than VGG16 in some instances. The original VGG architectures consist of five stacks of convolutional layers, each followed by max pooling layers. The top layers are the same across all VGG designs and consist of two fully-connected layers, each of size 4096 with 50\% dropout, and a fully-connected softmax layer of size 1000. 

\subsection{VGGFace}
Shortly after the release of the VGG architectures, the Visual Geometry Group published another paper called ``Deep Face Recognition" \cite{vggface}. In this paper, VGG16 is trained from scratch for facial recognition using a dataset of 2.6 million face images. Prior studies have shown that transfer learning using neural networks with facial recognition pretraining can produce highly effective results for gender recognition and age estimation \cite{effective_training}. Since facial recognition neural networks have already been trained to distinguish human features, the features that they extract may be more useful for determining age and gender from a photo than the features extracted by a more general neural network. In this study, VGGFace, VGG16 with facial recognition weights, is also examined for its proficiency at age and gender classification.

\section{The Datasets}
Machine learning models 
rely on the quality of data that feed them. Mislabeled data and excessive noise can cause models to start learning the wrong things. In deep learning, large and accurate datasets are essential to obtaining good performance. In this study, the MORPH-II dataset is used to train and test models.

\subsection{MORPH-II}
MORPH-II \cite{ricanek2006morph} is a good candidate for gender and age or other face image studies for a few reasons. The images captured are of the subjects' heads and most are positioned in front of a gray background -- which helps reduce background noise. Age labels and other information are provided about the subjects such as race, gender, and a unique identifier. A visual inspection of the images, however, reveals a few noisy variation. The subjects' heads are tilted in different directions, and may be of varying distance from the camera. Pixelations are apparent in most images, and some images have vastly different tint. 
The dataset consists of 55,134 images with subject ages ranging from 16 to 77 years old. 84.6\% of the dataset is male, and 77.22\% of the dataset is black. As seen in table \ref{age_by_gender}, few images exist of subjects 50 years of age and older. Because of this, a subsetting strategy has been adopted by the academic community from the works of Guo and Mu \cite{guo2011simultaneous}. They propose to divide the dataset into three subsets. The first two, $S_1$ and $S_2$, consist of only blacks and whites, and have a 3:1 male to female ratio. $S_3$ contains all of the remaining images. 

\begin{table}[htbp]
\caption{Age by Gender in MORPH-II}
\centering
\begin{tabular}{@{}p{0.04\textwidth}*{6}{L{\dimexpr0.19\textwidth-12\tabcolsep\relax}}@{}}
\toprule
& \textbf{$<$20} & \textbf{20-29} & \textbf{30-39} & \textbf{40-49} & \textbf{50+} & \textbf{Total} \\
\midrule
\textbf{Male} & 6649 & 14009 & 12436 & 10082 & 3468 & 46644 \\
\textbf{Female} & 836 & 2305 & 2924 & 1978 & 447 & 8490 \\
\bottomrule
\textbf{Total} & 7485 & 16314 & 15360 & 12060 & 3915 & 55134 \\
\end{tabular}
\label{age_by_gender}
\end{table}

\begin{figure}[thbp] \centering
\includegraphics[width = 90mm]{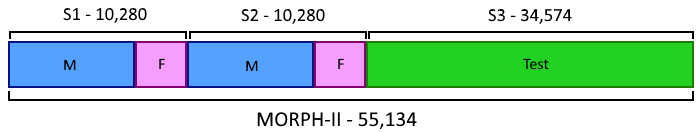} 
\caption{\small A depiction of subsetting scheme on a cleaned version of MORPH-II, following the subsetting proposed by Guo and Mu \cite{guo2011simultaneous}.}
\label{morph_sets}
\end{figure}

\subsection{MORPH-II Cleaned}
A survey of the MORPH-II dataset revealed several inconsistencies. Some subjects had different dates of birth. Others had multiple race labels or both gender labels. 
In order to combat the effects of ``dirty data", 
such inconsistency in age, gender and race has been manually identified and cleaned up for MORPH-II dataset. More details can be found in \cite{inconsistencies_and_cleaning}. Hereafter, the MORPH-II cleaned data are used in this project. Following the subsetting scheme proposed in \cite{guo2011simultaneous}, our subsetting scheme is shown in figure \ref{morph_sets}.  
Sets 1 and 2 both contain 10,280 images, and set 3 contains 34,344 images. 

\subsection{MORPH-II Equalized}
Preprocessed MORPH-II dataset are also considered in our preliminary study for performance evaluation. 
In this case, MORPH-II images are first cropped to fit the subjects' faces. During the process, the images are rotated such that the subjects' eyes are aligned. Images are grayscaled, and the lighting of the images is equalized. This dataset, known as MORPH-II equalized, is used in the early testing stages of our preliminary study due to its small input vector size \cite{image_preprocessing}. The full-sized images are either 200x240 or 400x480 having input vector sizes of 144,000 and 576,000 respectively, but the equalized images only produce 4,200 data points.

\section{Training Parameters}
\label{training_params}
To compare the effects of changes in transfer learning techniques, all training parameters are kept consistent unless otherwise specified. For MORPH-II, all images are scaled down to 200x240. All input is standardized before being fed into the network. The batch size is set to 50, and models are trained for 60 epochs. The original dropout rate of 0.5 is retained, and the ReLU activation function is used in all weight layers. The Adadelta optimizer is used with its default values. Gender models use the binary cross entropy loss function, and age models use mean absolute error (MAE). Results for age estimation are reported as an MAE, which is defined as: 
\begin{align}
{\text MAE} = \frac{1}{n}\sum_{i=1}^{n}|y_i - \hat{y_i}|. 
\end{align}
As such, MAE is the average of the absolute differences between the predicted age and the subject's actual age. For gender, the results are reported as an accuracy -- the number of correct predictions over the size of the test set. 

$S_1$ is used to train the models. During the parameter training process, models are supplied with a validation set of 500 random samples from $S_3$. To show the performance of models as data are added, the training set is split into several sets that are trained upon serially. This also helps avoid the issues that arise from using too much computer memory. The model parameters with the lowest loss on the validation set sample is saved and then fully validated on $S_2\cup S_3$, a set of 44,624 images.

\section{Top-Layer Retraining}
\label{top-layer_retraining}
\begin{figure}[htbp] \centering
\includegraphics[width = 90mm]{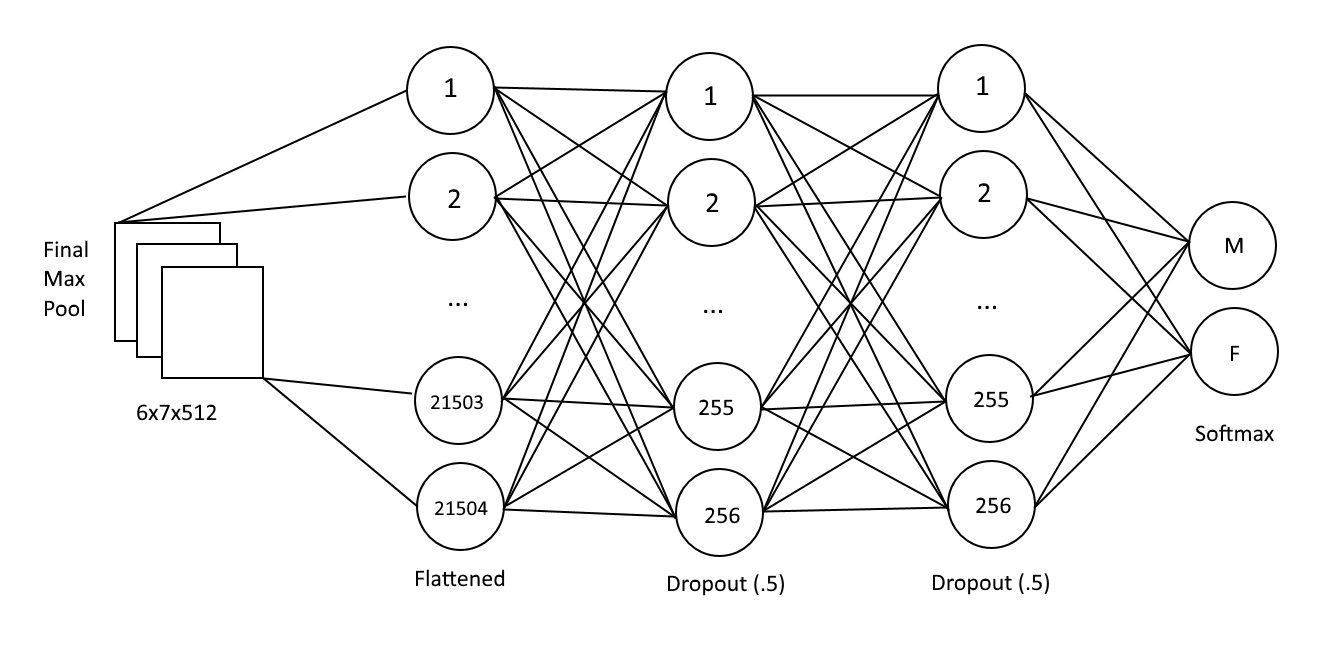} 
\caption{\small A new top for VGG19. All weights are initialized randomly and then the network learns how to discern male from female by examining the output of the convolutional layers.}
\label{top_layers}
\end{figure}

A common practice in transfer learning is to remove the top layers of a DNN, and then replace them with a different top. In VGG19, the top of the network is responsible for interpreting the output of the many underlying convolutional layers, so the same feature extractions are performed, but the new top layers produce predictions for the new task. The added top layers must be retrained from scratch, and are commonly initialized with random weights. During the training process, the rest of the network is frozen, so the weights in those layers maintain their initialized values and do not change during training. VGG19, for example, can be modified to be trained for gender recognition using the design shown in figure \ref{top_layers}. Note that in these experiments the dense layer sizes have been decreased to shorten training time and lower graphics card memory consumption. The output layer has been reduced to just two neurons -- one for a prediction of male and the other for a prediction of female. The ILSVRC (ImageNet Large Scale Vision Recognition Challenge) weights are frozen inside of VGG19 in the first 16 convolutional layers. Using a model's ImageNet weights is known as \textit{general task} pretraining, and can produce surprisingly good results on a wide variety of datasets \cite{NIPS2014_5347} \cite{effective_training} \cite{DBLP:journals/corr/XieJBLE15}. 

\subsection{Dense Layer Size} 
\begin{figure}[htbp] \centering
\includegraphics[width = 90mm]{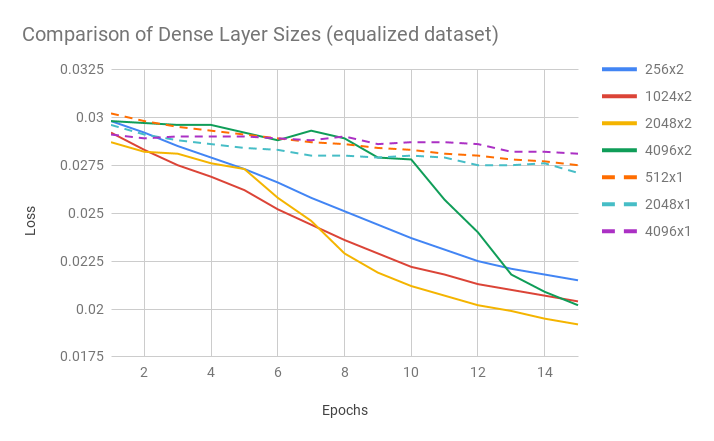} 
\caption{\small These are the losses on a short training test using different sizes and numbers of fully-connected layers. The models were trained on 20,000 images and validated on 5,000.}
\label{dense_layers}
\end{figure}

There are many factors that can be considered when retraining the top layers of a neural network. An obvious first choice is to test the size and number of fully-connected layers. As seen in figure \ref{dense_layers}, using only one dense layer seems to inhibit much of the learning process. Loss decreases more quickly when two dense layers are used. This gives traction the argument that more neurons will lead to a higher accuracy. A lower loss does not exactly equate to better validation results, but the general trend is that accuracy increases as loss decreases. The largest top layers, 2048x2 (two dense layers of 2048 neurons) and 4096x2, take several epochs before loss starts decreasing. This effect could be because of the randomly initialized weights. When the top layers are initialized with random weights, it takes more time for large layers to adjust and begin fitting the data. Much smaller dense layers, such as 16x2 and 32x2, also work for gender recognition with about 95\% accuracy. Age estimation requires larger dense layers than gender to produce good results.  

\subsection{Epochs}
Another commonly explored training parameter is the number of epochs for which to train a model. An \textit{epoch} is one pass through all of the data in the training set. Depending on the dataset, depth of the network, regularization techniques, and a variety of other factors, an optimal number of epochs might be high or low. Too few epochs and the network will be underlearned. Too many epochs and the model becomes overfit. In both of these instances, validation loss will be higher than usual, and it is unlikely that a near-optimal model will be produced. Figure \ref{epochs} below shows the $S_2\cup S_3$ test results at increasing numbers of epochs. For this test, the optimal number of epochs seems to be around 90 where an MAE of 4.753 is achieved. ArgMax and expected value are ways of decoding age from the softmax layer of the neural net and are explained in section \ref{argmax}.

\begin{figure}[htbp] \centering
\includegraphics[width = 90mm]{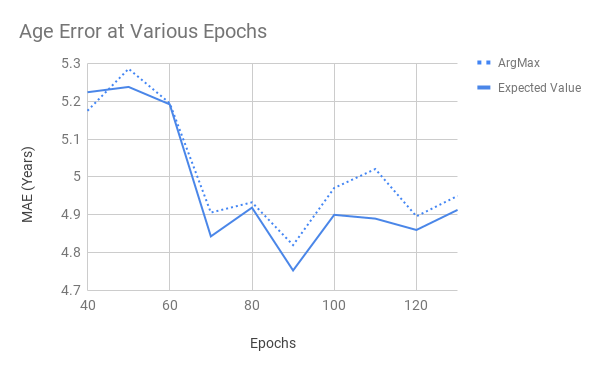} 
\caption{\small With epochs, there is a breaking point where a model goes from underlearned to overfit. Usually the best model occurs at that point.}
\label{epochs}
\end{figure}

\subsection{Dropout Regularization}
\begin{figure}[htbp] \centering
\includegraphics[width = 90mm]{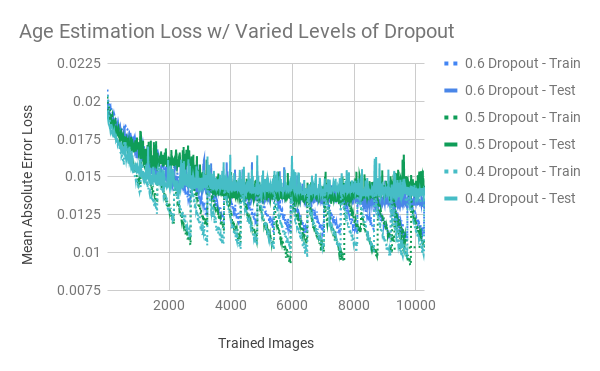} 
\caption{\small With lower dropout, the validation loss can be seen to improve more quickly, but it does not reach the depths of the losses that occur at a higher dropout rate. Especially towards the end of the training process, higher dropout can be seen achieving lower losses.}
\label{dropout}
\end{figure}

One method of combatting overfitting is to add dropout to weight layers. In the 2014 paper that introduces dropout, Srivastava et al. state that it ``provides a way of approximately combining exponentially many different neural network architectures efficiently" \cite{dropout}. When dropout is added to a weight layer, neurons are randomly selected to be removed from the network at each iteration. Those neurons are omitted both when the mini-batch is being fed through the network, and also during backpropagation. The number of neurons removed from the layer is determined by the dropout rate which is set manually. In VGG19, dropout is only used in the top fully-connected layers. Using a higher dropout rate provides more of a regularizing effect, but causes the model to not learn as quickly. In figure \ref{dropout}, dropout can be seen preventing overfitting as the training set losses stay higher, but the test set losses decrease. In table \ref{dropout_results}, the best and final models are also compared. During training, the model that achieves the lowest loss on the validation set is saved and is considered the ``best" model. After the last epoch of the last set of trained images, the final model is saved. The best model usually outperforms the final model except if the lowest loss happens to be obtained very early on in the training process. If the model overfits the validation set, a very good loss and accuracy might be recorded, but when fully tested, the performance is mediocre. In the dropout results table, the best result is seen with lower dropout because it acts like a model that is trained for more epochs than 60 which is optimal in this case. 

\begin{table}[htbp]
\caption{Dropout Results}
\centering
\begin{tabular}{@{}p{0.04\textwidth}*{5}{L{\dimexpr0.19\textwidth-12\tabcolsep\relax}}@{}}
 & \textbf{0.3} & \textbf{0.4} & \textbf{0.5} & \textbf{0.6} & \textbf{0.7}\\
\midrule
\textbf{Best} & \textbf{4.768} & 5.002 & 5.192 & 4.915 & 5.056 \\
\textbf{Final} & 4.953 & 4.949 & 6.065 & 5.698 & 5.143\\
\midrule
\end{tabular}
\label{dropout_results}
\end{table}

\section{Training Techniques}
\label{training_techniques}
In addition to changes in the network, many different training techniques were compared in order to observe their effects on gender recognition and age estimation. Training techniques are ways of training a model that can result in better accuracy. Improvements due to the training techniques explored in this section do not result from changes in the network, but from changes in the data. 

\subsection{Input Standardization}
\begin{figure}[htbp] \centering
\includegraphics[width = 90mm]{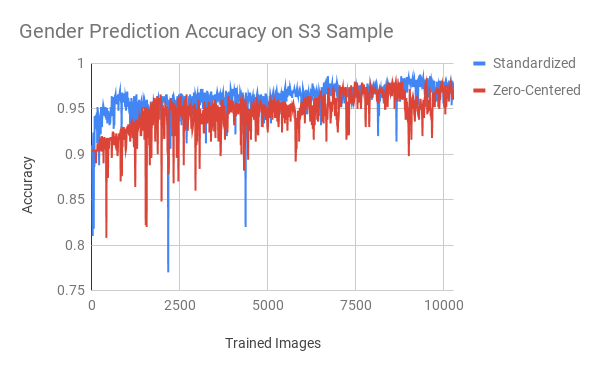} 
\caption{\small Standardizing the input helps VGG19 reach higher accuracy faster.}
\label{standardization}
\end{figure}

When VGG was originally submitted to the ImageNet competition, the creators trained on images that had been \textit{zero-centered}. This means that the average was calculated for the training set and subtracted from each pixel value before being fed into the network. Like zero-centering, standardizing image data also centers it at zero, but additionally gives the pixel values a normal spread. When standardizing training data, the validation and test data must also be standardized with respect to the training data. The mean $\bar{x}$, and standard deviation $\sigma$, of $S_1$ can be seen in table \ref{avg_sd}. Once they have been calculated, the formula:
\begin{align}
\frac{P_i - \bar{x}}{\sigma}
\end{align}
can be applied to $P$ which is the set of all pixel values (red, green, and blue) in $S_1$. As seen in figure \ref{standardization}, standardizing the input data produces immediately better results. Not only does the model begin to fit the dataset faster, it also reaches a higher accuracy than the zero-centered dataset. The accuracies shown in figure \ref{standardization} are validation set accuracy during each epoch of training. Once trained, the standardized model reaches a gender classification accuracy of 96.209\% on the full $S_2\cup S_3$ test set. This is 1.083\% higher than the performance of the model trained on a zero-centered $S_1$.

\begin{table}[htbp]
\caption{$S_1$ Mean and Standard Deviation}
\centering
\begin{tabular}{@{}p{0.04\textwidth}*{3}{L{\dimexpr0.19\textwidth-12\tabcolsep\relax}}@{}}
 & $\bar{x}$ & $\sigma$ \\
\midrule
\textbf{$S_1$} & 142.46 & 59.85 \\
\midrule
\end{tabular}
\label{avg_sd}
\end{table}

\subsection{Data Augmentation}
\begin{figure}[htbp] \centering
\includegraphics[width = 90mm]{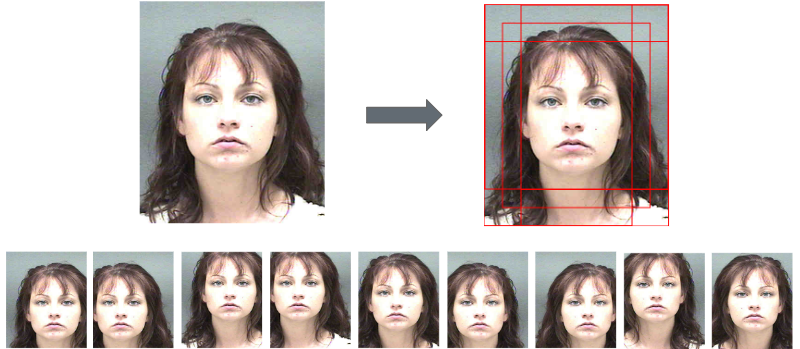} 
\caption{\small The original 200x240 images are cropped and flipped to become ten 160x200 images.}
\label{cropping}
\end{figure}

Data augmentation techniques are commonly used to train neural networks \cite{alexnet} \cite{googlenet}. Since large and accurate datasets are rare and usually private, data augmentation can be used to create more data with which to train a network. For this project, 12-crop resampling was tested. This involves taking a crop from each corner of the image and the center of the image, and resizing the image down to crop size. These six samples are then flipped horizontally to produce twelve unique images. Figure \ref{cropping} shows an example of a MORPH-II image after 10-crop resampling has been applied (12-crop resampling without the resized image). Using this technique and the training parameters described in section \ref{training_params}, an MAE of 5.028 years was obtained. This is a slight increase in performance over the 5.192 MAE of the baseline test, but a drawback is that the network must be trained on 123k 160x200 images instead of 10k 200x240 images, taking 8 times as long to train. 

\subsection{More Data} \label{more_data}
Many suggest that a large dataset is integral to deep learning. A sales pitch for deep learning is that deep neural networks can learn more from the data and thereby surpass traditional statistical methods. To test the effects of a larger dataset, the training data are doubled in size by making use of $S_2$. $S_1 \cup S_2$ becomes a set of 20,560 images and is trained upon, while $S_3$ is used for testing. The sales pitch appears to ring true as the model achieves an MAE of 4.690 years. Part of the drop in MAE is due to the drop in female population. In $S_2 \cup S_3$, females make up 13.1\% of the population but in $S_3$ the female population is 9.6\%.

\subsection{Label Distribution Age Encoding (LDAE)}
LDAE is a method of encoding age that has proven more effective than simple one-hot encoding \cite{effective_training}. LDAE recognizes that people age differently, so it helps to view a person's age, denoted by $A$,  as a small scope of potential ages rather than just a binary truth. In this method, the formula:
\begin{align}
\label{ldae_formula}
f(i|A, \alpha) = \frac{1}{ \sqrt{2\pi} \alpha}e^{-\frac{(i-A)^2}{2\alpha ^ 2}}
\end{align}
is used to calculate a probability at each age to encode age labels. In the formula $A$ is the age label, $i$ is the age for which a probability should be produced, and $\alpha$ is a hyperparameter that affects the spread of the age probabilities. 

\subsubsection{ArgMax and Expected Value}
\label{argmax}
Two ways that an age can be decoded from the output of a neural network are known as ArgMax and expected value. ArgMax uses the age that has the highest probability. Expected value multiplies the probability at each age by the age and then sums the products. In most cases, expected value gives more accurate predictions, but the age MAEs are usually fairly close. 

\subsubsection{Dynamic LDAE}
In general, it is easier to mistake an old person's age by a large amount than a young person's age.
To represent the differing certainty between young and old, the $\alpha$ value in Equation \ref{ldae_formula} is increased linearly as age increases. 
In this paper, dynamic LDAE is proposed as follows: an overall $\alpha$ of 2.5 is considered, with a higher $\alpha$ for old ages and a lower $\alpha$ for young ages. 
To illustrate: During training, age labels are encoded with LDAE from the ages of 5 to 85. The resulting input vector has 81 dimensions, each containing a probability for the corresponding age. For example, considering $\alpha$ with a range from 1 to 4, and for age $A$, the dynamic  $\alpha$ is estimated by:
\begin{align}
\label{alpha_formula}
g(\alpha|A) = \frac{range(\alpha)} {81} * (A-5) + 1.
\end{align}

\begin{figure}[htbp] \centering
\includegraphics[width = 90mm]{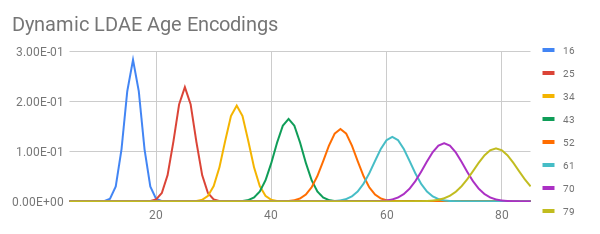} 
\caption{\small Training image age labels are encoded with dynamic LDAE. Eight age encodings are shown as $\alpha$ ranges from 1 to 4 and age ranges 16 to 79.}
\label{ldae}
\end{figure}

An illustration of the input vector encodings can be seen in figure \ref{ldae}. It  improves the accuracy as seen in table \ref{ldae_results}.

\begin{table}[htbp]
\caption{Dynamic LDAE Results}
\label{ldae_results}
\centering
\begin{tabular}{@{}p{0.08\textwidth}*{4}{L{\dimexpr0.22\textwidth-13\tabcolsep\relax}}@{}}
 & One-Hot & $\alpha$ = 2.5 & $\alpha$ = 1-3.5 & $\alpha$ = 1-4 \\
\midrule
\textbf{MAE} & 5.250 & 5.192 & 4.861 & \textbf{4.778} \\
\midrule
\end{tabular}
\end{table}

\section{Results with VGGFace}
\label{vggface_results}
All of the tests from sections \ref{top-layer_retraining} and \ref{training_techniques} use VGG19 with its ILSVRC weights. This section considers transfer learning with the VGG16 architecture pretrained for facial recognition. Although VGG19 is capable of detecting the subtlest differences to separate millions of images into 1000 classes, some of the filters it has learned activate on mundane objects or animal fur, so they do not produce strong activations on images containing human faces. Since VGGFace was originally trained to separate several captures of 2662 individual faces \cite{vggface}, every filter in VGGFace is geared towards finding human facial features. The models in this section were trained using most of the same parameters as above and no extra data augmentation or generalization techniques. As can be seen in tables \ref{vggface_mae} and \ref{vggface_gender}, VGGFace takes far fewer epochs to fit the training data even though it is a smaller network. Gender validation set accuracy, for example, reaches 98\% during the first epoch of $S_1$. Because of this, smaller increments of epochs must be searched to find the best resulting models. All models in this section are still trained on $S_1$ and tested on $S_2 \cup S_3$. As expected, VGGFace produces better results than a general task VGG19 network.

\begin{table}[htbp]
\caption{VGGFace Expected Value MAE}
\centering
\begin{tabular}{@{}p{0.05\textwidth}*{7}{L{\dimexpr0.19\textwidth-13\tabcolsep\relax}}@{}}
\toprule
\textbf{Epoch} & \textbf{5} & \textbf{10} & \textbf{15} & \textbf{20} & \textbf{25} & \textbf{30} & \textbf{35}\\
\midrule
\textbf{MAE} & 4.800 & 4.483 & 4.443 & 4.468 & \textbf{4.322} & 4.377 & \textbf{4.323} \\
\bottomrule
\end{tabular}
\label{vggface_mae}
\end{table}

\begin{table}[htbp]
\caption{VGGFace Gender Prediction Accuracy}
\centering
\begin{tabular}{@{}p{0.05\textwidth}*{7}{L{\dimexpr0.19\textwidth-13\tabcolsep\relax}}@{}}
\toprule
\textbf{Epoch} & \textbf{3} & \textbf{6} & \textbf{9} & \textbf{12} & \textbf{15} & \textbf{18} & \textbf{21} \\
\midrule
\textbf{Accy.} & 98.59\% & 98.47\% & \textbf{98.68}\% & 98.53\% & 98.65\% & 98.64\% & 98.56\% \\
\bottomrule
\end{tabular}
\label{vggface_gender}
\end{table}

\section{Gender-Specified Hierarchal Age Model}

In the past, hierarchies of classifiers have used multiple feature labels of datasets to increase classification accuracy. Test data are separated into different classes before being classified again by models trained specifically for each class. Guo and Mu use a hierarchy of KPLS (kernel partial least squares) race and gender classifiers with BIFs to obtain a MORPH-II MAE of 4.18 years \cite{guo2011simultaneous}. It makes sense that men and women would have different features that are indicative of age, so a hierarchy of deep CNN models might also produce better age estimation results than a single model used for both genders. Because female image data are fairly limited in MORPH-II, an 80/20 train/test split was devised such that no subjects who appear in the training data are also in the testing data. A gender model was trained for the hierarchy using the same 80\% portion of the training data and it achieved an accuracy of 98.60\%.

\begin{figure}[htbp] \centering
\includegraphics[width = 60mm]{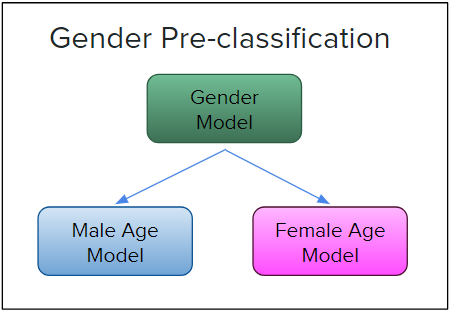} 
\caption{\small In this design, images are classified as male or female before estimating age with the corresponding model.}
\label{preclassifier}
\end{figure}

The female age model achieved an MAE of 5.22 years after being trained for 25 epochs on all female images in the training set. The male age model achieved an MAE of 3.79 years after 20 epochs. The deep CNN hierarchy achieved an MAE of 4.10 years which outperforms all other age estimation models in this document. This experiment shows that with enough training data, a hierarchy of deep CNNs can surpass a single model trained for general age estimation. Additionally, the model can make age and gender predictions at a rate of 62.62 frames per second meaning that it is suitable for real-time age and gender estimation deployment. Table \ref{comparison} shows a comparison of the results from this paper alongside other MORPH-II studies.

\begin{table}[htbp]
\caption{MORPH-II Results with Comparison}
\centering
\begin{tabular}{@{}p{.055\textwidth}*{6}{L{\dimexpr0.19\textwidth-11\tabcolsep\relax}}@{}}
\toprule
\textbf{Approach} & \textbf{Year} & \textbf{Train} & \textbf{Test} & \textbf{Age} & \textbf{Gender} \\
\midrule
BIF+OLPP \cite{guo2010human} & 2010 & $S_1$, $S_2$ & $S_2 \cup S_3$, $S_1 \cup S_3$ & 4.45 & 97.84\% \\
BIF+KPLS \cite{guo2011simultaneous} & 2011 & $S_1$, $S_2$ & $S_2 \cup S_3$, $S_1 \cup S_3$ & 4.18 & 98.20\% \\
VGG19 & 2018 & $S_1$ & $S_2 \cup S_3$ & 4.75 & 96.6\% \\
VGGFace & 2018 & $S_1$ & $S_2 \cup S_3$ & 4.32 & 98.68\% \\
Gender-Specified & 2018 & 80\% & 20\% & 4.10 & 98.60\% \\
\bottomrule
\end{tabular}
\label{comparison}
\end{table}

\section{Conclusion}
Although VGG19 was not originally trained to recognize faces, good results for gender recognition and age estimation can still be obtained using transfer learning techniques. Transfer learning with a pretrained model that is more pertinent to the task, such as VGGFace, can produce results that beat other gender recognition and age estimation techniques, and can even exceed human performance. Changes in network designs and training techniques can be studied without having to spend weeks training models from scratch. The models for this paper were all trained using a GTX 1060 Max-Q and a GTX 1070. 
 This paper has shown the advantages offered by certain model designs, training techniques, and pretrained weights. It has also demonstrated that hierarchies of AI models offer promise and should be considered when implementing a classification system. 

{\bf Future Work} The results shown here surpass nearly all results obtained before 2012 simply by leveraging new deep learning technology. They are, however, far from what is possible. A live demo of these AI models revealed flaws. Age estimations were occasionally wildly off for minorities in the dataset such as Asians, Hispanics, women, children, and elderly people. Additionally, gender predictions seemed largely based on the absence or presence of long hair and could change by the tilt of the head. To address these types of issues further training techniques could be studied that help to generalize and stabilize prediction accuracy. Larger and more balanced datasets could be used as training data. Newer types of deep CNN architectures could be adapted to age and gender estimation and might yield better results even with general task weights. Increasingly deep hierarchies of models can be considered for appropriately large datasets. Combinations of datasets and fusions of features could be what's on the horizon for the advancement of deep learning. Finally, training models from scratch specifically for age and gender predictions would probably produce better results.

\section{Acknowledgements}

This research was conducted with funding from the National Science Foundation under DMS Grant Number 1659288. 
{\small
\bibliographystyle{IEEEtran}
\bibliography{references.bib}
}
\end{document}